\documentclass[letterpaper, 10 pt, conference]{ieeeconf}  

\IEEEoverridecommandlockouts                              

\usepackage{graphics} 
\usepackage{epsfig} 
\usepackage{mathptmx} 
\usepackage{times} 
\usepackage{amsmath} 
\usepackage{amssymb}  
\usepackage[ruled]{algorithm2e}
\usepackage{subfigure}
\usepackage{tabu}
\usepackage{booktabs}
\usepackage{multirow}
\usepackage{footnote}
\usepackage{threeparttable}
\usepackage{booktabs}
\usepackage[table]{xcolor}
\usepackage[letterpaper,top=60pt,bottom=43pt,left=48pt,right=48pt]{geometry}

\title{\LARGE \bf
	Federated Imitation Learning: A Novel Framework for Cloud Robotic Systems with Heterogeneous Sensor Data
}

\author{Boyi Liu$^{{1},{3}}$, Lujia Wang$^{1}$, Ming Liu$^{2}$ and Cheng-Zhong Xu$^{4}$
	\thanks{*This paper was recommended for publication by Okamura, Allison upon
		evaluation of the Associate Editor and Reviewers' comments. This research is supported by the Shenzhen Science and Technology Innovation Commission (Grant Number JCYJ2017081853518789), the Guangdong Science and Technology Plan Guangdong-Hong Kong Cooperation Innovation Platform (Grant Number 2018B050502009) and the National Natural Science Foundation of China (Grant Number 61603376) awarded to Dr. Lujia Wang. The National Natural Science Foundation of China (Grant Number U1713211) ,the Shenzhen Science, Technology and Innovation Commission (Grant Number JCYJ20160428154842603) and Basic Research Project of Shanghai Science and Technology Commission (Grant No. 16JC1401200) awarded to Prof. Ming Liu.}
	\thanks{$^{1}$Boyi liu, Lujia Wang are with Cloud Computing Lab of Shenzhen Institutes of Advanced Technology, Chinese Academy of Sciences. {\tt\small liuboyi17@mails.ucas.edu.cn};
		{\tt\small lj.wang1@siat.ac.cn}}
	\thanks{$^{2}$Ming liu is with Department of ECE, Hong Kong University of Science and Technology. {\tt\small eelium@ust.hk}}
	\thanks{$^{3}$Boyi liu is also with the University of Chinese Academy of Sciences.}
	\thanks{$^{4}$Cheng-Zhong Xu is with the University of Macau.  {\tt\small czxu@um.edu.mo}}
}

\begin{document}
	\maketitle
	\thispagestyle{empty}
	\pagestyle{empty}
	
	\begin{abstract}
		Humans are capable of learning a new behavior by observing others to perform the skill. Similarly, robots can also implement this by imitation learning. Furthermore, if with external guidance, humans can  master the new behavior more efficiently. So, how can robots achieve this? To address the issue, we present a novel framework named FIL. It provides a heterogeneous knowledge fusion mechanism for cloud robotic systems. Then, a knowledge fusion algorithm in FIL is proposed. It enables the cloud to fuse heterogeneous knowledge from local robots and generate guide models for robots with service requests. After that, we introduce a knowledge transfer scheme to facilitate local robots acquiring knowledge from the cloud. With FIL, a robot is capable of utilizing knowledge from other robots to increase its imitation learning in accuracy and efficiency. Compared with transfer learning and meta-learning, FIL is more suitable to be deployed in cloud robotic systems. Finally, we conduct experiments of a self-driving task for robots (cars). The experimental results demonstrate that the shared model generated by FIL increases imitation learning efficiency of local robots in cloud robotic systems.
	\end{abstract}

	\section{INTRODUCTION}
	In tradition imitation learning scenarios, demonstrations provide a descriptive medium for specifying robotic tasks. Prior work has shown that robots can acquire a range of complex skills through demonstrations, such as table tennis [1], drawer opening [2], and multi-stage manipulation tasks [3]. Nevertheless, there exist a number of problems in the application of imitation learning. For example, large amounts of data are required and sometimes they are heterogeneous. These drawbacks result in long training time for the robot and limited generalization performance. Cloud robotic system [4] can be adopted to increase the “learning efficiency” of robots, and federated imitation
		\begin{figure}[thpb]
		\centering
		\includegraphics[width=1\linewidth]{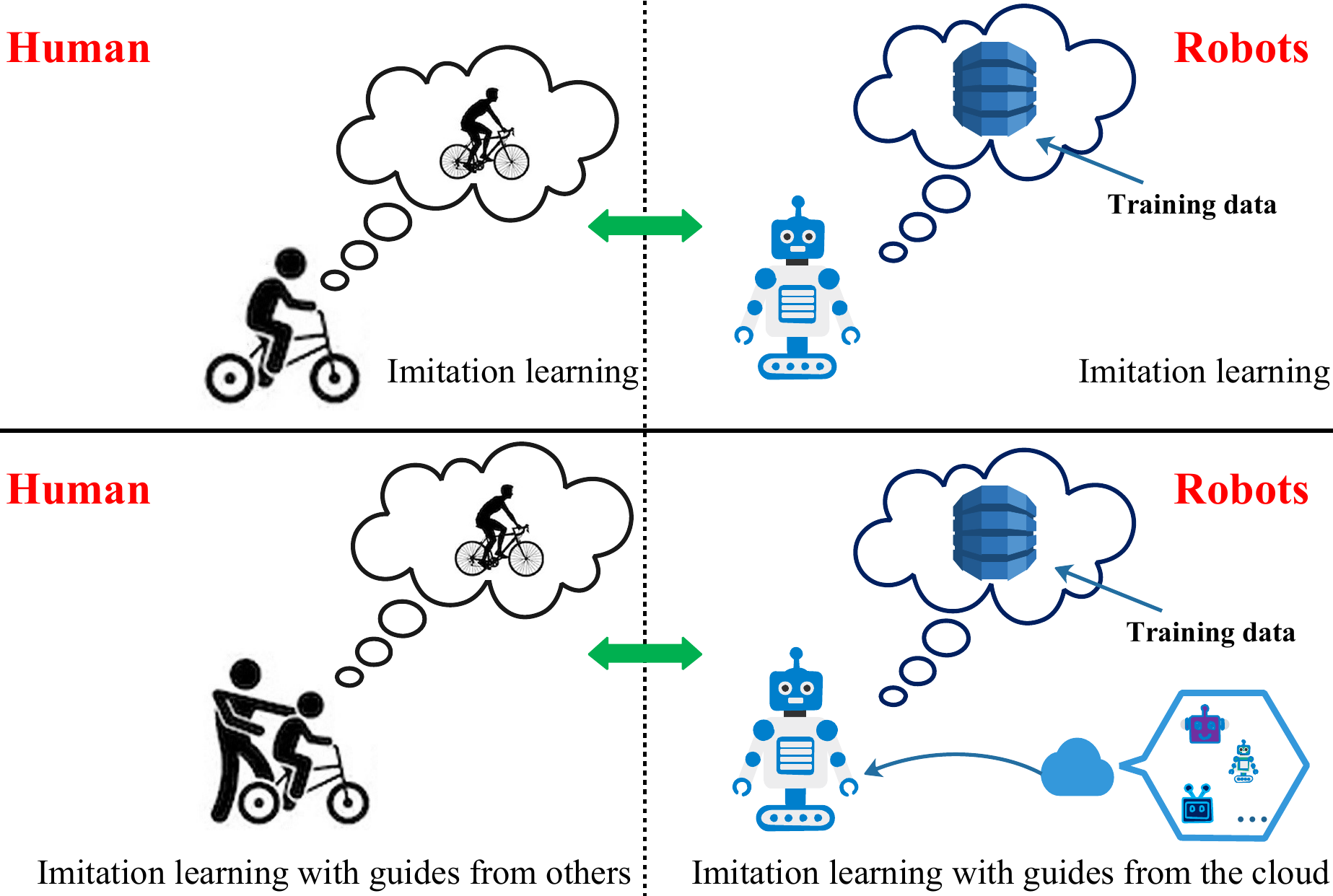}
		\caption{The child on the upper left acquires the ability to ride a bicycle by observing an adult. This is the process of imitation learning in humans. Correspondingly, the upper right robot acquires skills by training data. This is the process of imitation learning in robots. The bottom left child not only acquires bicycling skills by observing an adult, but also gets helps from an adult. This makes his learning more efficient. Inspired by this, in this work, FIL enables the bottom right robot not only acquires skills by training data, but also gets knowledge from other robots through the cloud robotic system.}
		\label{fig:architecture}
	\end{figure} 
	learning algorithm is proposed to fuse the shared knowledge of robots.

	The cloud fuses knowledge which is from local robots. However, data heterogeneity hinders the process. This issue is generally regarded as a major challenge of cloud robotic systems [5]. To overcome this issue, a novel framework named FIL has been proposed. It increases imitation learning of local robots in heterogeneous data condition. As shown in Fig.1, it is inspired by the case that humans can learn more effectively if they have external guidance. With FIL, a robot is capable of taking advantage of knowledge from other robots. The student/teacher imitation learning in local robots will be incrseased with the guide model provided by FIL. To evaluate it, we conduct an autonomous driving task. Experimental result indicates that FIL enables robots to absorb knowledge from other robots and apply the knowledge to increase imitation learning efficiency and accuracy. Videos of the results can be found on the supplementary website\footnote[1]{The video is available at https://sites.google.com/view/federated-imitation}. Overall, this paper makes the following contributions:
	\begin{itemize}
		\item We present a novel framework named FIL. It provides a knowledge fusion mechanism for cloud robotic systems.
		\item We propose a knowledge fusion algorithm in FIL. It enables the cloud to fuse knowledge from local robots and generate guide models for robots with service requests.  
		\item Based on transfer learning, we present a knowledge transfer scheme to facilitate local robots acquiring knowledge from the cloud.
	\end{itemize}
	\section{Related Work}
	In this work, FIL is proposed to increase the learning efficiency of each local robot in the cloud robotic system. The cloud generates a shared model and provides guidance services for each local robot. Methods that can achieve similar goals include transfer learning and meta-learning. Compared with these two approaches, FIL achieves heterogeneous knowledge fusion without raw data sharing. The related work are introduced as follows.
	\subsection{Transfer learning}
	Transfer learning aims at improving the performance of target learners on target domains by transferring the knowledge contained in different but related source domains, which is similar to the goal of FIL. In recent years, transfer learning research communities are mainly focused on deep transfer learning [6]. The techniques used in deep transfer learning include four categories: instances-based transfer learning, mapping-based transfer learning, network-based transfer learning, and adversarial based transfer learning [7]. 
	
	The above approaches have made efforts for knowledge transferring between domains. Unfortunately, the difference between domains in a cloud robotic system can be large. Sometimes the data are collected with different kinds of sensors. Therefore, the datasets of local robots in the cloud robotic system may be heterogeneous. It is not possible to directly use the transfer learning technology. Therefore, the proposed framework first fuses heterogeneous knowledge  rather than using transfer learning technology directly.
	\subsection{Meta-learning}
	Meta-learning and the proposed framework have the same ultimate aim. Meta-learning, or learning to learn, is the science of systematically observing how different machine learning approaches perform on a wide range of learning tasks, and then learning from this experience, or meta-data, to learn new tasks much faster than otherwise possible [8]. Applications of meta-learning in robotics have achieved good results. Among these, One-Shot Visual Imitation Learning [9] and Domain-Adaptive Meta Learning (DAML) [10] by Abbeel's lab are representative. Compared with the former that can enable robots to perform imitation learning from action videos of robots, DAML is an improved approach, which enables the robot to imitate human actions directly. DAML enables a robot learning to visually recognize and manipulate a new object after observing just one video demonstration from a human user. To enable this, DAML uses a meta-training phase where it acquires a rich prior over human imitation, using both human and robot demonstrations involving other objects. DAML extends a prior meta-learning approach to allow for learning cross domain correspondences and includes a temporal adaptation loss function. 

	However, the phase of meta-training is essential in DAML or other meta-learning approaches. It means that we have to obtain data of all local robots because the step of calculating loss to update gradients requires all task data (corresponding local data in the cloud robotic system). It is an impossible task considering limited communication. The more effect of meta-learning requires more data and a higher-capacity model to support. Nevertheless, it is difficult for the cloud to get all the data of local robots. Therefore, meta-learning approaches are unsuitable for the cloud robotic system in this work, although they may get higher accuracy.
	\subsection{Knowledge sharing in cloud robotic system}
	Knowledge sharing is the key in the work, which means local robots share knowldge with each other. For this purpose,  [11] presented a mental simulation service for the existing OPENEASE cloud engine. With this service available, agents can delegate computationally-expensive simulations to more powerful computing services. In addition, they created a SWI-Prolog library so that the developers and robots can describe the world state, self abilities and the problem. [12] proposed a framework for robots to reason on a central knowledge base which has ontologies and execution logs from other robots. This approach achieves knowledge exchanging with cloud robotic.   [13] presented a platform that supports large-scale on-demand data collection for robot. All of the above three approaches can realize the fusion of local knowledge. Unfortunately, they can't fuse the knowledge from heterogeneous data, which is addressed by this work. 
	\section{Methodology}
	In this section, we will present the details of the proposed framework and algorithms, which includes knowledge acquiring technology, framework of FIL, knowledge fusion algorithm and knowledge transferring algorithm. 
	\subsection{Knowledge acquiring by imitation learning}
	Local robots acquire knowledge through imitation learning in FIL. Imitation learning is commonly posed either as behavioural cloning [14] or as inverse reinforcement learning [15], both of which require demonstrations. Imitation learning has empowered recent advances in learning robotic manipulation tasks by addressing shortcomings of reinforcement learning such as exploration [16] and reward specification [17]. The knowledge acquiring approach used in FIL of local robots belongs to behavioural cloning, which focuses on learning the expert’s policy using supervised learning. The way behavioural cloning works is quite simple. Given demonstrations of robots, we will divide these into state-action pairs. We treat these pairs as i.i.d. examples and finally, we apply supervised learning.
	
	\subsection{Framework of FIL}
	The framework of FIL is performed in Cloud-Robot-Environment setup. There are local robots, cloud servers, communication services and computing device. Local robots learn skills through imitation learning and the cloud server fuses knowledge. We develop a federated learning algorithm to fuse private models into the shared model in the cloud. With the shared model, the cloud server is capable of generating guide models corresponding to requests of local robots. After that, the local robots perform transfer learning based on the guide model. Finally, the final policy will be quickly obtained. As illustrated in Fig.2, we will explain the methodology in FIL with the example of a self-driving task.
	\begin{figure}[thpb]
		\centering
		\includegraphics[width=1\linewidth]{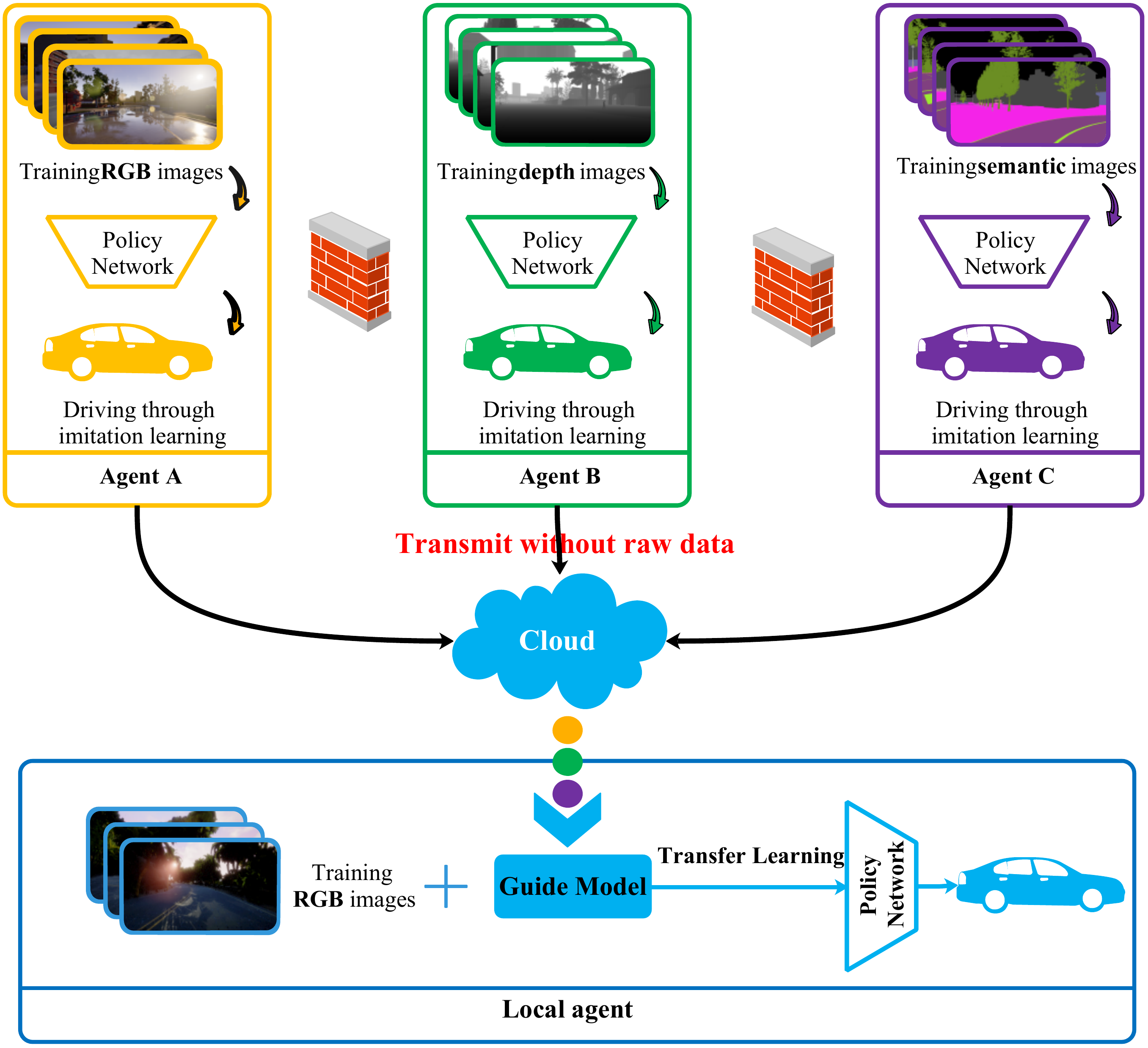}
		\caption{FIL framework. The work assumes that there are three agents to fulfil the task. They perform imitation learning to acquire the policy models with heterogeneous data: RGB images, depth images and semantic segmentation images. Neither can the raw training data be shared between agents nor between agents and clouds. Only the parameters of the policy models will be uploaded and fused, and then the cloud will provide guide models while robots request.}
		\label{fig:architecture}
	\end{figure}
	
	Based on the self-driving task, we typically collected three types of data. The three agents use three different types of dataset separately. Datasets of local robots are labeled but will not be sent to the cloud. Three different policy models will obtained by local training. RGB images will be trained by Agent A, and a private policy model (Private Model A) will be obtained. Actions of robots will be determined by the output which might be some actions or  parameters. Similar processes occur in agent B and agent C. Outputs of the three private models are with same types. The inputs to each model are different: RGB images, depth images, and semantic segmentation images. Then the parameters of all three models will be uploaded to the cloud and fused there. Henceforth, the cloud will be capable of generating guide models for different types of input. When a local robot requests a service, the cloud will provide a guide model in correspondence with the type of sensor data. FIL can be performed either online or offline. As presented in Algorithm 1, the whole framework can be summarized as following steps:
	\begin{itemize}
		\item Step1: Imitation learning performed by local robots;
		\item Step2: Parameters (of private models) Transmitting;  
		\item Step3: Fusing (in the cloud) knowledge;
		\item Step4: Responding to the local requests and generating guide models for them.
	\end{itemize} 
	Noted that step1 and step 2 are simultaneous while FIL performs online. Labels of the cloud data will be updated simultaneously.
	\begin{algorithm}
		\caption{Processing Algorithm in FIL}
		Initialize action-value Q-network with random weights ${\theta }$;
		
		\KwIn{$n$: number of local robots ; $q$: update frequency.} 
		\While{cloud server is running}
		{d
			${\theta }_{nt}\leftarrow {robot }_{n}$ performs imitation learning
			
			\If{$t\%q==0$}
			{
				\For{$i=0;i<n;i++$} 
				{ 	
					Send ${\theta }_{it}$ to the cloud;
				} 
				labels=fuse(${\theta }_{1t}$, ${\theta }_{2t}$,$\cdots$,${\theta }_{nt}$)
			}
			\If{service\_request=True}
			{
				Generate ${\theta }_{cloud}$ base on labels;
				Send ${\theta }_{cloud}$ to local robots;
				
				${\theta }_{local}$=transfer(${\theta }_{cloud}$).
			}
		}
	\end{algorithm}
	\begin{figure*}[thpb]
		\centering
		\includegraphics[width=1\linewidth]{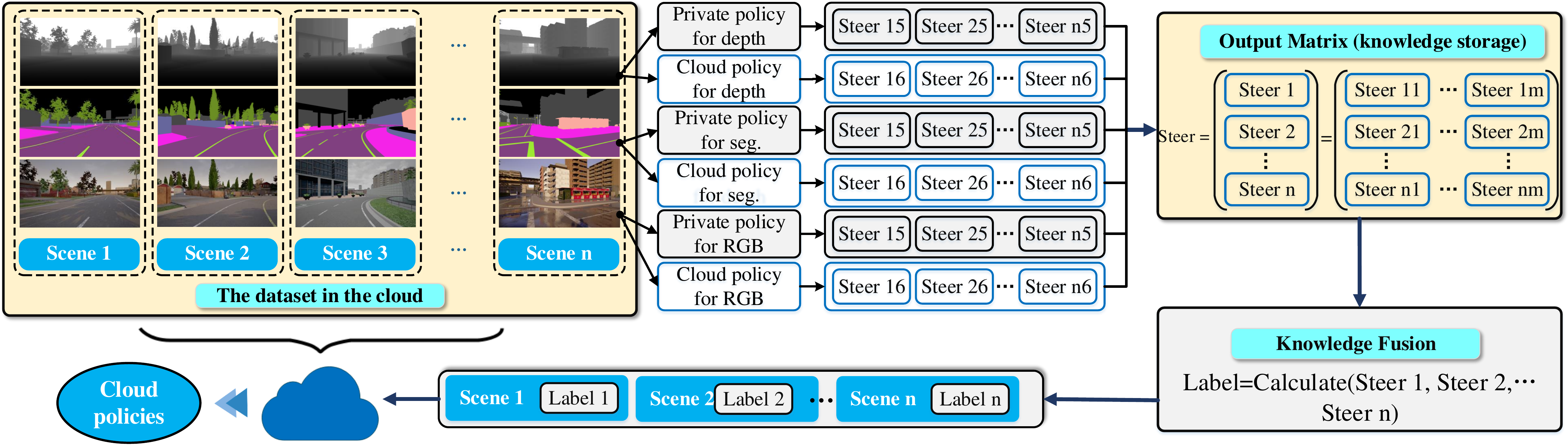}
		\caption{Knowledge fusion algorithm deployed on the cloud in a self-driving case. Agents obtain private models by performing imitation learning. The cloud stored different types of sensor data of many scenes. Input corresponding sensor data in the cloud dataset to private models. Then calculate the numerical characteristics of private models outputs to label scene. Noted that these data is not uploaded from local but collected by the cloud, and the cloud dataset is much larger than each local dataset. With multi types of sensor data, the cloud is capable of generating guide models corresponding to the sensor type of the local robot.}
		\label{fig:architecture}
	\end{figure*}
	\subsection{Knowledge fusion algorithm in the cloud}
	Knowledge fusion has been a focus of many approaches since the end of the 20th century. These studies focus on the construction of knowledge bases [18] [19] or knowledge representations [20] [21]. However, the above approaches that define the knowledge representation of the local robots are unsuitable for cloud robotic systems. We cannot determine how local robots obtain data and express knowledge. In our work, the proposed framework aims to make local robot learn efficiently and smart. In another word, it is one of the demonstrations of robot lifelong learning.
	
	Existing machine learning approaches fuse knowledge by centralizing storage of training data. Consequently, these approaches are unattainable in cloud robotic systems with large scale local datasets considering the limited communications. In [22], the federated learning system was introduced. Therein, the mobile devices perform computation of model training locally on their training data according to the model released by the model owner. Such design enables mobile device to collaboratively learn a shared model while keeping all the training data on the device. Similarly, multi-sensor data fusion is required by many robotic tasks, such as the navigation of self-driving cars, mobile robot SLAM. But it is impossible to upload all kinds of sensor data to the cloud. Therefore, we propose a federated imitation learning framework in cloud robotic systems to improve the ability of robots. In the proposed framework and algorithm, there is no need to upload raw sensor data to the cloud, only the parameters of the models are shared in the cloud.
	
	Fig.3 presents the knowledge fusion algorithm. Primary responsibility of the cloud is to label scenes. Before this, some data collecting for cloud training is necessary, and the types of these data should at least cover the types of local datasets. For example, in the self-driving task of this example, the RGB images, depth images and semantic segmentation images should be collected for the cloud. Thus, one scene has three types of sensor data. Each has a corresponding uploaded private model which provide its own suggestion of labelling to the cloud. And then, these suggestions will be congregated to produce a final label for this scene. The calculation approach of labelling can draw on some methods of ensemble learning which can be defined according to different application scenarios, in this case, we choose the median of outputs. As private models will output the steer of the agent. Agents usually make decisions of following the road, turning, obstacle avoiding, etc. Generally speaking, the output of the model includes two extreme cases: turning and no turning, where errors occur most. While Median can avoid extreme values in the evaluation. For example, if there were 5 local models and outputs of them for one scene are: -0.1, -0.3, 0.4, 0.4, 0.5. We will take 0.4 as the label of current scene. As data being labeled, cloud models will be trained immediately. Formula (1) to Formula (5) have summarized the whole process:
	\begin{equation}
	R_{{D_i}}^{emp}(\theta)=\frac{1}{N} \sum_{n=1}^{N}{L}\left({{y}_i}^{(n)}, f\left({{x}_i}^{(n)} ; \theta\right)\right)
	\end{equation}
	In the Formula (1), ${D_i}=\left\{\left(\mathbf{x}^{(n)}, y^{(n)}\right)\right\}_{n=1}^{N}$ is the dataset of the local robot i. L represents the loss function.$\theta$ repensents parameters of models. $x_i$ are original data and $y_i$ are labels. N is the number of samples of the dataset. $R^{emp}_{D_i}$ represents the empirical risk in training sets.
	\begin{equation} 
	\theta_{i}^{*} =\underset{\theta}{\arg \min } {R}_{{D}}^{struct}(\theta)
	\end{equation}
	The Formula (2) presents training targets of local robots. It is the structure risk minimization criteria. $R^{struct}_{D}$ represents structure risk in datasets.
	\begin{equation} 
	l_{in} =f_{{\theta}i}\left(scene_{in}\right)
	\end{equation}
	In the above formula, $scene_{in}$ is the training data in the cloud. i represents sensor types, n represents the number.
	\begin{equation} 
	Me_{in} =Median\left(l_{in}\right)
	\end{equation}
	\begin{equation} 
	\theta_{i(cloud)}^{*} =\underset{\theta}{\arg \min } \frac{1}{M} \sum_{n=1}^{M} {L}\left(Me_{in}, f\left(scene_{in}^{(n)} ; \theta\right)\right)+\frac{1}{2} \lambda\|\theta\|^{2}
	\end{equation}
	In Formula (4) and (5), $Me_{in}$ is the median of $l_{in}$, $\theta_{i(cloud)}^{*}$ is the training targets in the cloud. M represents the number of sample data. $\theta$ is the regularization term of L2 norm, which is used to reduce the parameter space and avoid over-fitting,  lambda is used to control the intensity of regularization. y is true labels in the dataset. l means the prediction from the model we trained. "in" means the n-th scene of the i-th type of data.
	
	Noted that we only use the shared model in the cloud as a guide model for local robots. The shared model maintained in the cloud is a cautious policy model, which means it will not make serious mistakes in some private unstructured environments but the action might not be the best. Thus, it is necessary for every local robot to train its own policy model based on the shared model received from the cloud. This is the transfer learning process in FIL.
	\subsection{Transfer the shared model}
	There have been a lot of valuable studies on transfer learning. Current research mainly focuses on transferring through the relationship between source domain distribution and target domain distribution. This method is unsuitable for cloud robotic systems because it requires raw data of local robots. Under the constraints of the above conditions, Layer Transfer is the transferable learning method that can be implemented. Layer Transfer means that some layers in the model trained by source data are copied directly and the remaining layers are trained by target data. The advantage of this is that target data only needs fewer parameters to be trained, thus avoiding over-fitting. It has faster training speed and higher accuracy. On different tasks, the layers that need to be transferred are often different. For example, in speech recognition, we usually copy the last layers and retrain the first layers. This is because the first layers of speech recognition neural network are the way to recognize the speaker's pronunciation, and the last layers are the recognition. The latter layers have nothing to do with the speaker. In image recognition, we usually copy the front layers and retrain the back layers. This is because the first layers of the image recognition neural network are to identify whether there is a basic geometric structure, so it can be transferred. The latter layers are often abstract and cannot be transferred. So, which layers to be transferred are case by case.
	\begin{figure}[thpb]
		\centering
		\includegraphics[width=0.75\linewidth]{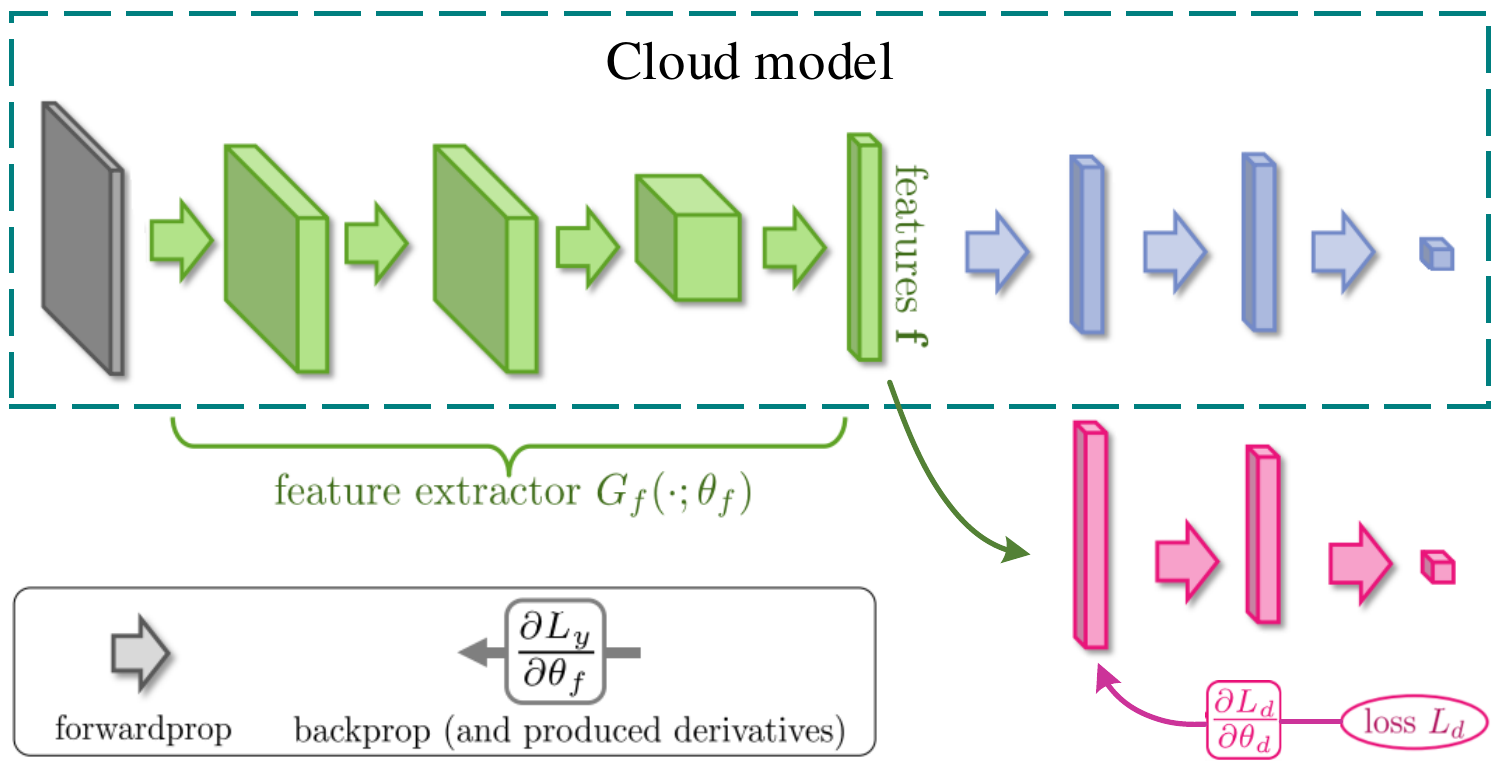}
		\caption{A transfer learning approach of FIL}
		\label{fig:architecture}
	\end{figure}
	
	As presented in Fig.4, in the work, we use front layers as feature extractors in the case of imitation learning. The decision model in the cloud can be used as the initial model for local training. In this way, cloud model can play a guiding role. It can speed up local training and increase the accuracy of local robots. In the training of local robots, the feature extraction layer is frozen and only the full connection layers are trained. If necessary, it can also adjust the relevant parameters in the process of back propagation. For example, some work may increase learning rate. After transfer learning, local robots successfully utilize knowledge from other robots in cloud robotic systems.
	
	\section{Experiments}
	In this section, we will present our experimental setup and the results. To verify the effectiveness of FIL, we have to answer two questions: 1) Is FIL capable of generating an effective shared model based on shared knowledge in cloud robotic systems? 2) have the shared model of FIL improved the learning process or accuracy of local robots? To answer the first question, we have conducted experiments to generate cloud models and compare its performance with general models. To answer the second question, we then have conducted experiments to compare the learning process and accuracy of local robots with FIL and without FIL.
	\subsection{Experimental setup}
	Self-driving car can be regarded as an advanced robot. So it is enough to use cars to verify robot control algorithms. In the work, we have used Microsoft AirSim and CARLA as our simulator to evaluate the presented approach. In addition to have high-quality environments with realistic vehicle physics, AirSim and CARLA have a python API which allows for easy data collection and control. 
	\begin{figure}[thpb]
		\centering
		\subfigure[Data collection for local]{
			\label{fig:subfig:a} 
			\includegraphics[height=1.18in]{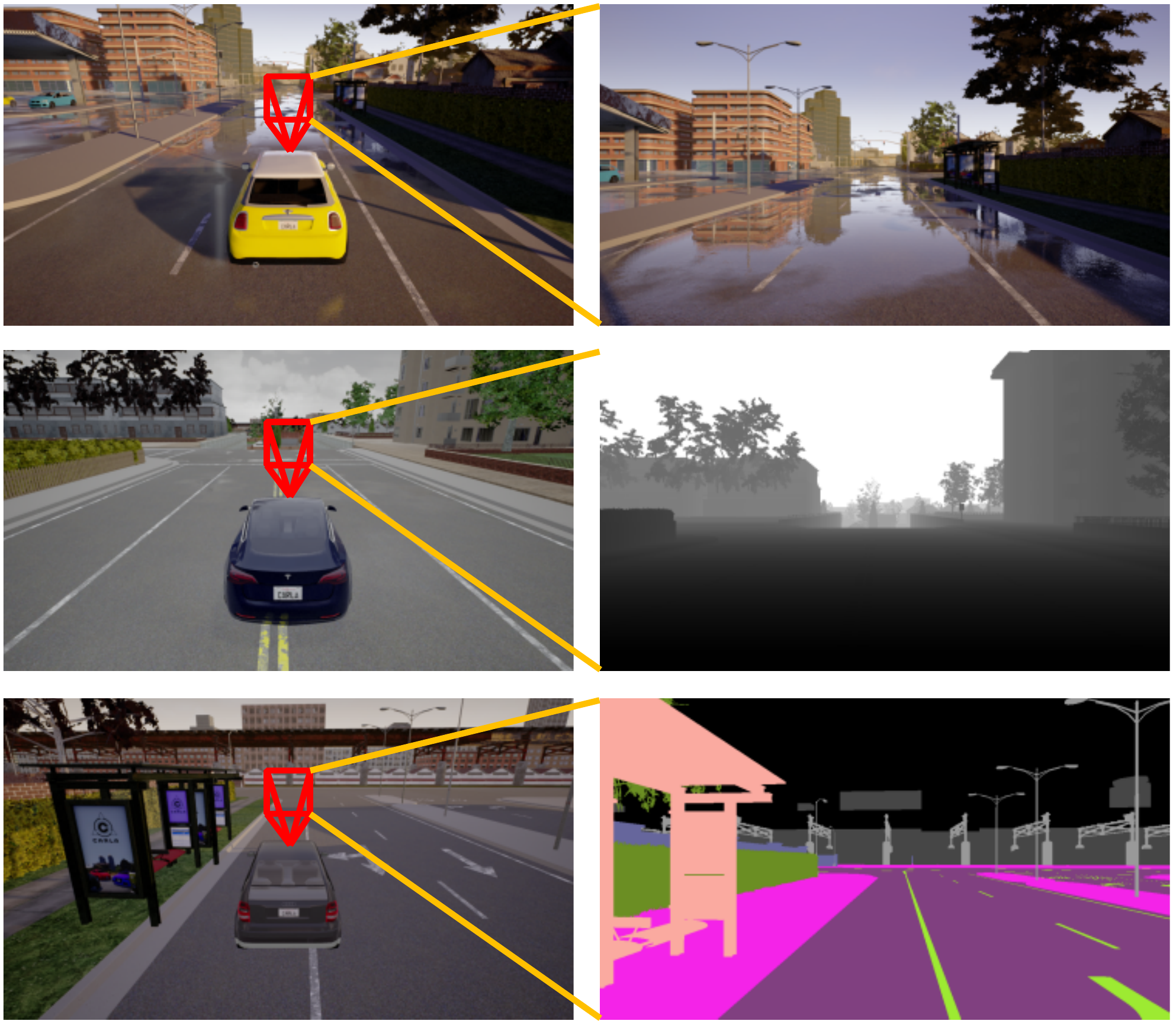}}
		\subfigure[Data collection for the cloud]{
			\label{fig:subfig:b} 
			\includegraphics[height=1.18in]{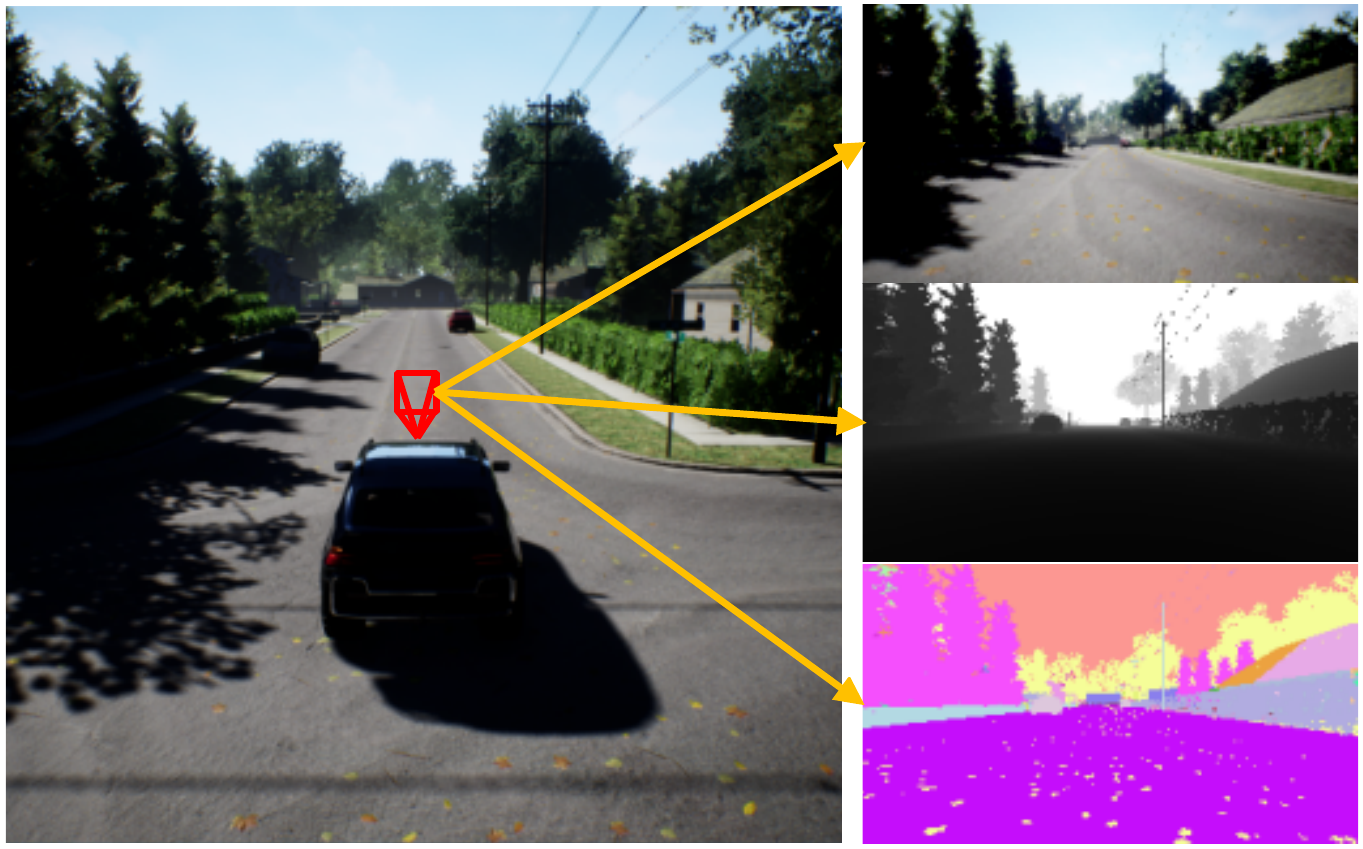}}
		\caption{As presented in subfigure (a), we collected training data for local robots in three different environments corresponding to three different types of data. As presented in subfigure (b), we collected different types but simultaneous data in many different environments.}
		\label{fig:subfig} 
	\end{figure}
	
	In order to collect training data, a human driver is presented with a first-person view of the environment (central camera). The simulated vehicle is controled using the keyboard by the driver. The car should be kept at a speed around 6m/s, collisions with cars or pedestrians should strive to be avoided, but traffic lights and stop signs will not be considered. As Fig.5 presents, we have used three different types of sensor data: RGB images, depth images and semantic segmentation images. Any of these three types of sensor data can make the agent to perform obstacle avoidance tasks within tolerable errors. The policy network mainly consists of convolution layers and fully connected layers. Then a linear layer followed by a softmax, and the value function by a linear layer. Its architecture presented in Fig. 6 is similar to VGG-16.
	\begin{figure}[thpb]
		\centering
		\includegraphics[width=1\linewidth]{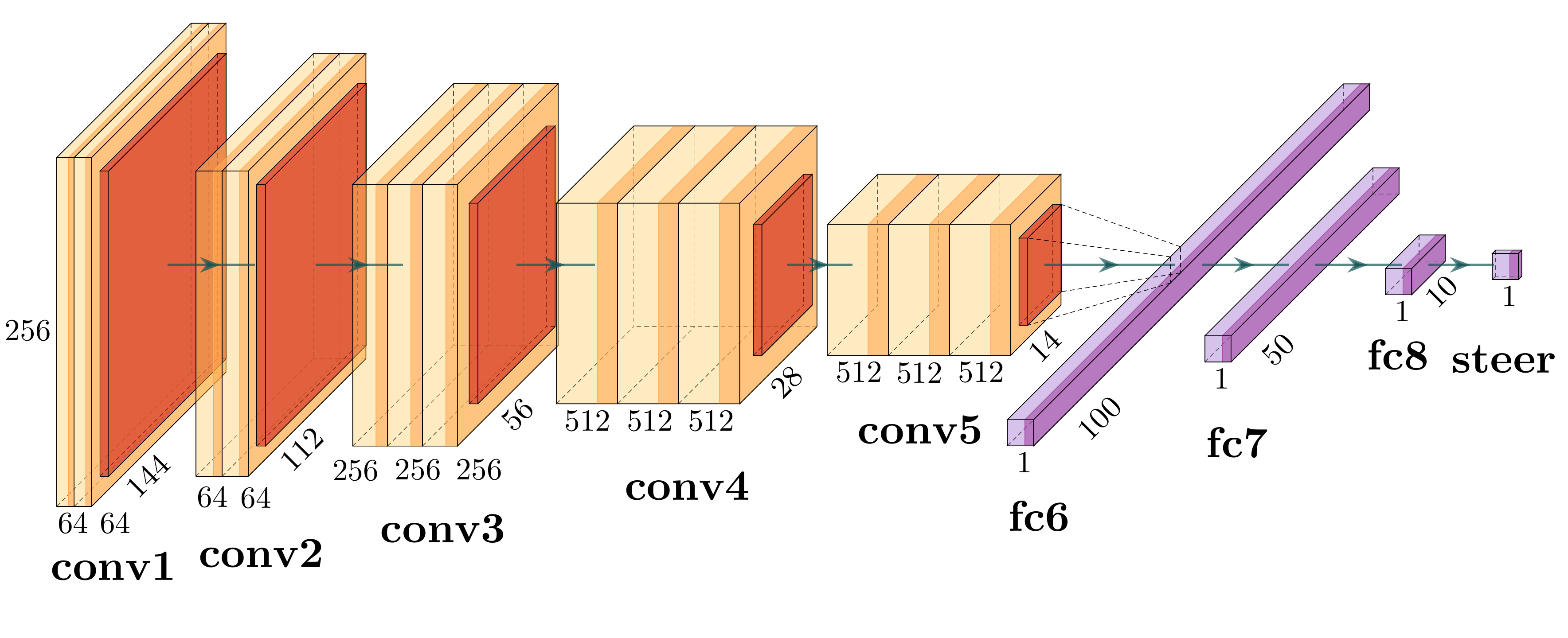}
		\caption{Network architecture of imitation learning in experiments}
		\label{fig:architecture}
	\end{figure}
	
	As for the cloud robotic systems, the server and agent communicate via HTTP requests, with the data server utilized the Django framework to respond to asynchronous requests. Our cloud programs run on the Microsoft Azure cloud. We conduct local robot experiments with a single NVIDIA Quadro RTX 6000, which allows us to run our simulator to receive photo-realistic images for training. 
	\begin{figure*}[thpb]
		\centering
		\includegraphics[width=1\linewidth]{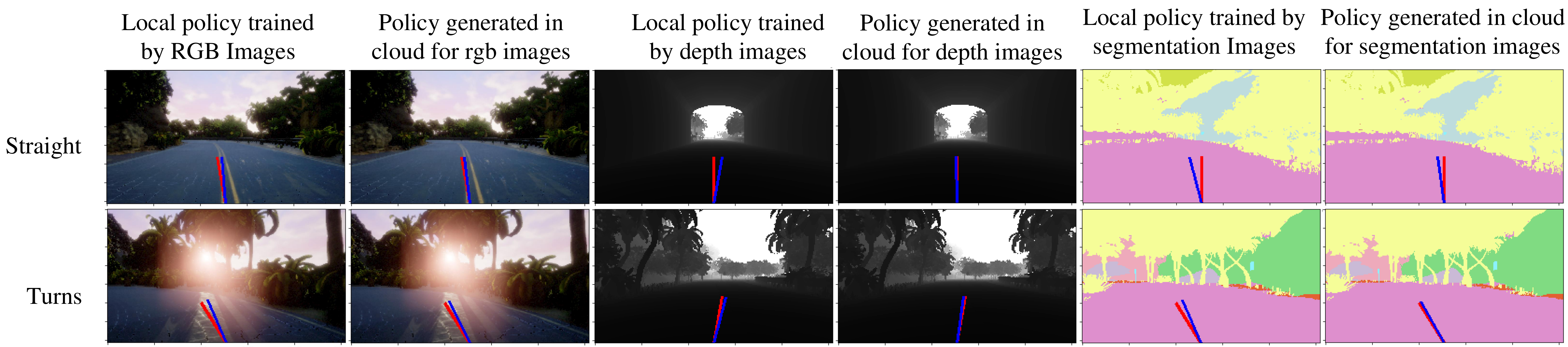}
		\caption{Performance comparison of shared models and original local models. The red lines indicate the steering angles output by the policy model. The blue lines indicate the steering angles of human driving. Performance of models are mainly reflected in turning and going straight. There are two rows in the table, the first row is performance of each model in turning task, the second row is the performance of each model in going straight task. The agents are not trained separately in these two tasks because the output of the policy model is the steering angle rather than actions.}
		\label{fig:architecture}
	\end{figure*}
	\subsection{Evaluation for the shared-model generating method in FIL}
	
	In this section, the robot (car) will challenge the tasks such as avoiding collisions and making timely turns. The observations (images) are recorded by one central camera. The recorded control signal is the steering angle. The steering angle is scaled between -1 and 1, with extreme values corresponding to full left and full right, respectively. Considering the actual driving, we transfer the steering angle between -0.69 radians and 0.69 radians.
	
	Once the three private policy networks been trained, the cloud will work to fuse their knowledge. As mentioned before, we assume that there are three companies train their policies by imitation learning with heterogeneous sensor data. Sharing the training data between agents or sending the raw data to the cloud is forbidden. So, the cloud server only gets the parameters of the three local networks and performs the knowledge fusion algorithm. Different types of sensor data is collected simultaneously from different environments in CARLA and Airsim before that. Then every scene in cloud will be labelled. The cloud generates local policy network 1 for company 1 based on RGB images, local policy network 2 for company 2 based on semantic segmentation images, and local policy network 3 for company 3 based on depth images. Then the parameters of these three private networks will be uploaded to the cloud. Finally, the cloud gets labels of datasets in the cloud. It will generate policy networks corresponding to different sensor requests. In this experiment, three policy networks will be generated. They are named cloud policies. The process in the cloud is unsupervised learning. There is no manual labels but cloud generation. For evaluation, we labeled some scenes to mark the result. In the simulation environment, the controller uses the policy network to control the robot.
	\begin{table}[htbp]
		\centering
		\caption{Results of local policies and cloud policies.}
		\setlength{\tabcolsep}{1mm}{
			\begin{tabular}{cccc}
				\toprule
				Controller & \multicolumn{1}{p{3.3em}}{Hit the obstacle} & \multicolumn{1}{p{3.3em}}{Miss turns} & \multicolumn{1}{p{4.2em}}{Mistakes in straight} \\
				\midrule
				Local controller for RGB images & 3.45\% & 12\%  & 16.67\% \\
				\rowcolor[rgb]{ .851,  .851,  .851}Cloud controller for RGB images & \textbf{0.69}\% & \textbf{0}     & \textbf{0} \\
				Local controller for depth images & 0     & 20\%  & 0 \\
				\rowcolor[rgb]{ .851,  .851,  .851}Cloud controller for depth images & \textbf{0}     & \textbf{4}\%   & \textbf{0} \\
				Local controller for segmentation images & 0     & 12\%  & 6.67\% \\
				\rowcolor[rgb]{ .851,  .851,  .851}Cloud controller for segmentation images & \textbf{0}     & \textbf{4}\%   & \textbf{0} \\
				\bottomrule
		\end{tabular}}%
		\label{tab:addlabel}%
	\end{table}%
	
	Fig. 7 presents results of these six policy networks in some main challenging scenes to the car. From Fig.7, it is clear that the cloud model presents higher accuracy than local models. So that it can avoid errors of local models training from single training set collected by one type of sensors. We conducted 3 experiments, each one sets a different starting point. Then we evaluated the performance of robots in obstacle avoidance, turning and straight forward tasks. The results are summarized in Table 1. It can be seen from the experimental results that the cloud knowledge improves the local controller that is trained using general imitation learning. The controller based on cloud policies performs better. Especially the controller for RGB images.
	\subsection{Evaluation for the knowledge-transfer ability of FIL}
	\begin{figure}[thpb]
		\centering
		\includegraphics[width=1\linewidth]{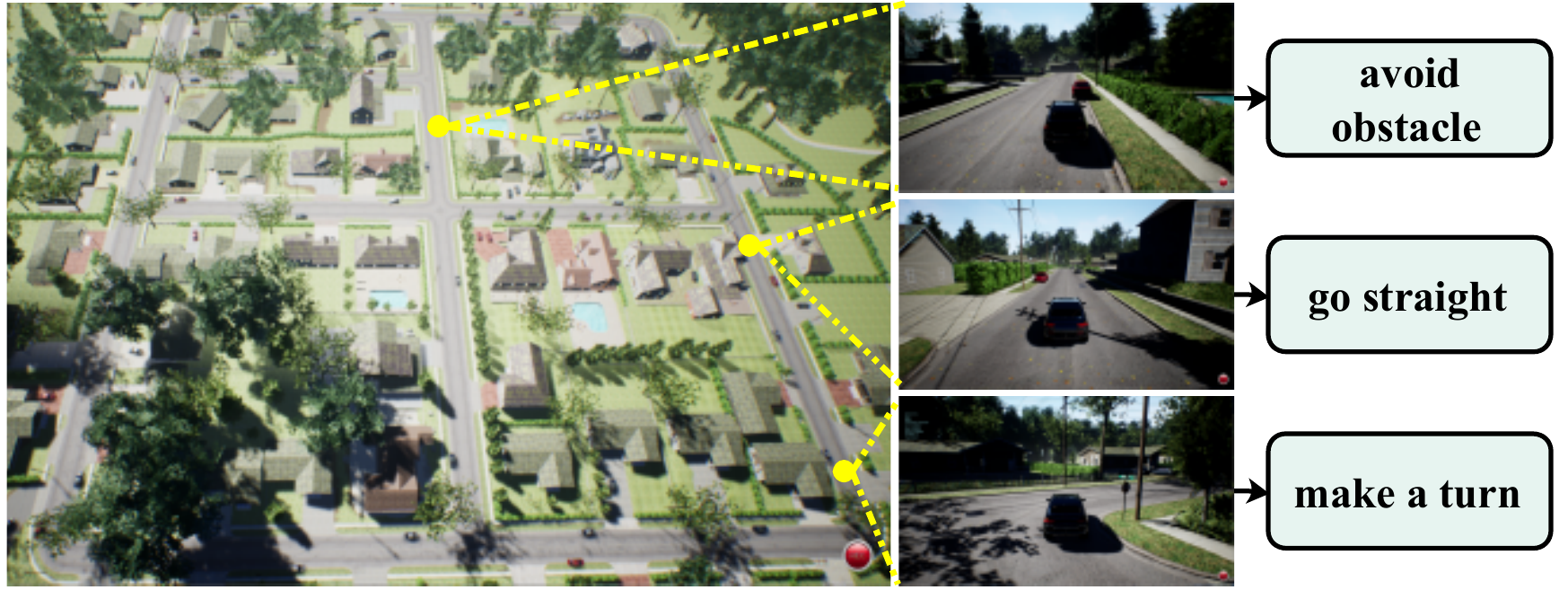}
		\caption{Challenges in the self-driving task}
		\label{fig:architecture}
	\end{figure}
	\begin{figure*}[thpb]
		\centering
		\includegraphics[width=1\linewidth]{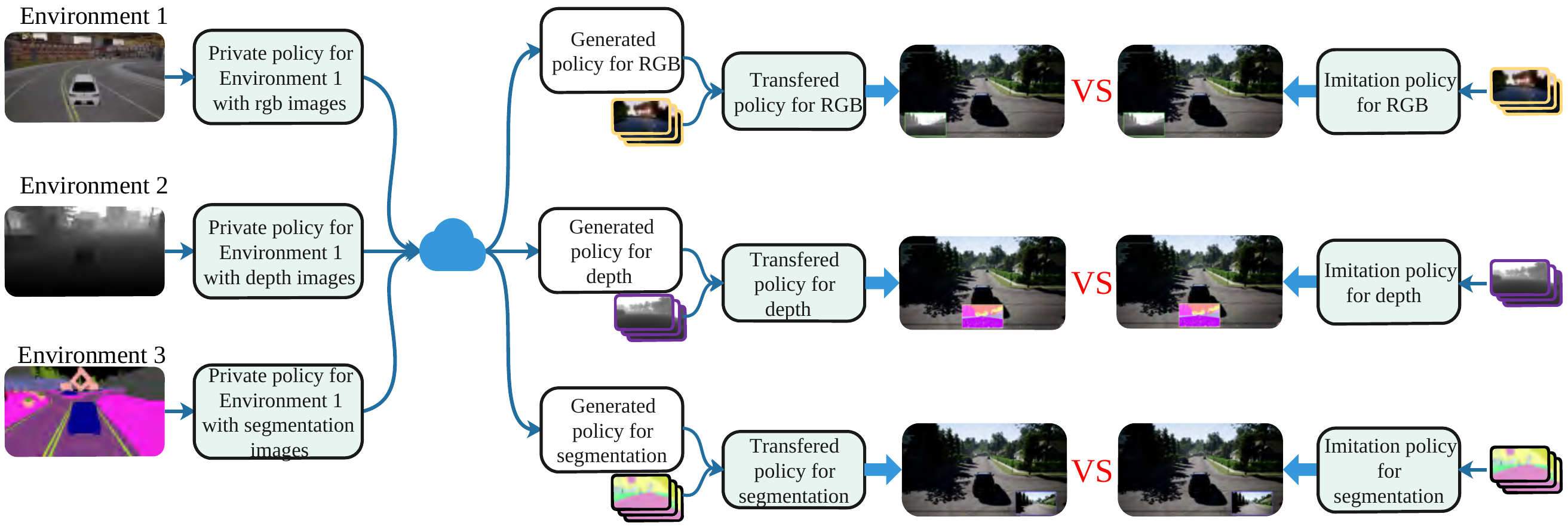}
		\caption{Procedure of the experiment for evaluating knowledge-transfer ability. Firstly, we performed imitation learning in three different environments with three types of sensor data (RGB images, depth images and semantic segmentation images) and got three policies. Secondly, the parameters of three private policies were sent to the cloud. The cloud generated labels of its own dataset. Here is something to explained is that there are semantic segmentation difference between CARLA and Airsim, so there is a unifying process for semantic segmentation images. Thirdly, three policies corresponding to different sensor data were generated. After that, we performed transferring learning with corresponded type of images and obtained transferred policies. For comparison, we also conduct the general imitation learning experiment on the right part of the image. Finally, we compare performances of these policies in a new environment (neighborhood in Airsim). }
		\label{fig:architecture}
	\end{figure*}
	\begin{figure*}[thpb]
		\centering
		\includegraphics[width=1\linewidth]{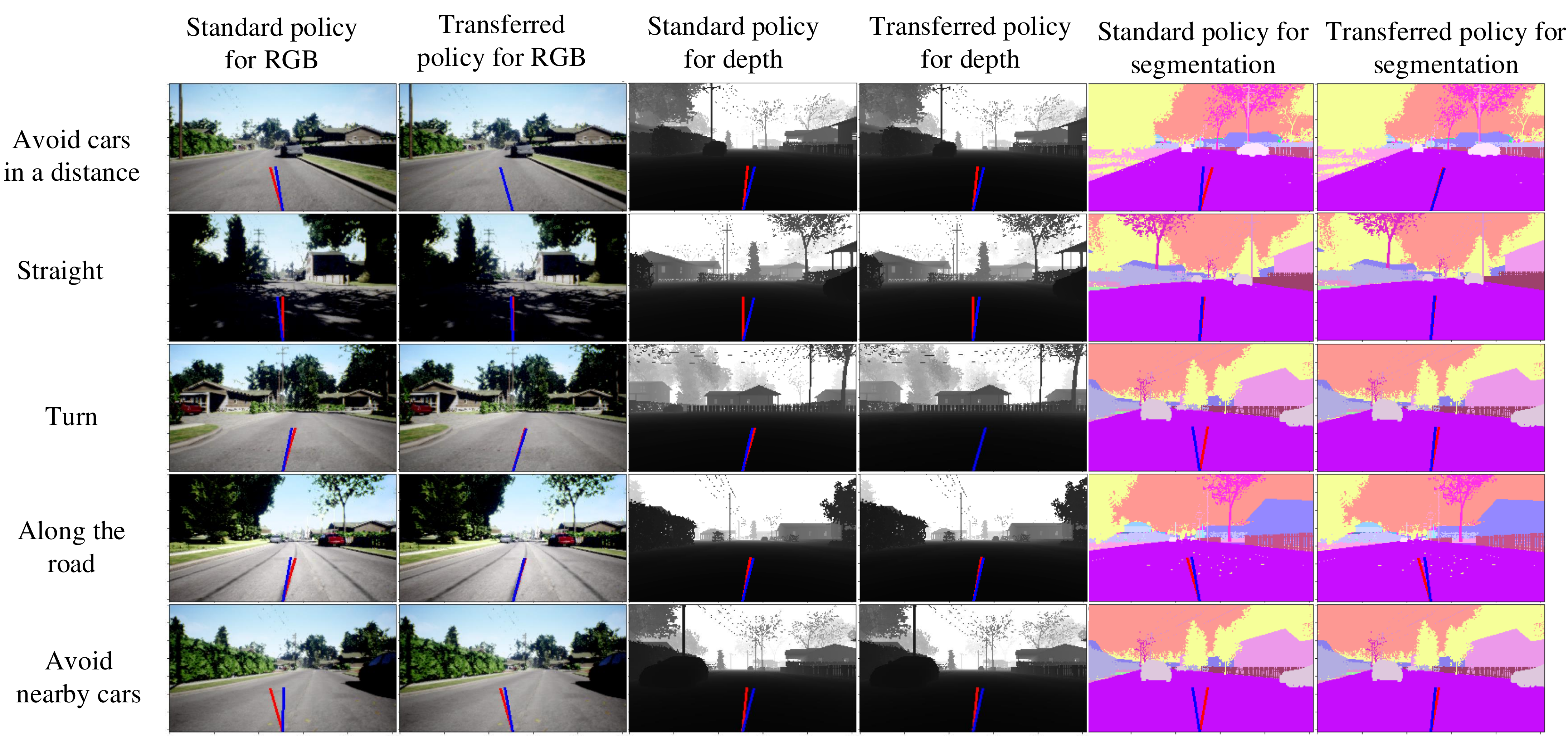}
		\caption{Performance comparison of general policies and transferred policies. The red lines indicate the steering angles output by the general policies. The blue lines indicate the steering angles of human driving. }
		\label{fig:architecture}
	\end{figure*}
	\begin{table*}[htbp]
		\centering
		\caption{Performance comparison of standard controllers and transferred controllers in different bad weather conditions}
		\begin{tabular}{cccccc}
			\toprule
			Controller & Error rate in normal & Error rate in rain & Error rate in snow & Error rate in fog & Error rate in dust \\
			\midrule
			Standard controller for RGB images & 17.39\% & 26.09\% & 30.43\% & 34.78\% & 52.17\% \\
			\rowcolor[rgb]{ .851,  .851,  .851} Transferred controller for RGB images & \textbf{4.35\%} & \textbf{8.70\%} & \textbf{17.39\%} & \textbf{31.82\%} & \textbf{39.13\%} \\
			Standard controller for depth images & 13.04\% & 4.35\% & 8.70\% & 8.70\% & 17.39\% \\
			\rowcolor[rgb]{ .851,  .851,  .851} Transferred controller for depth images & \textbf{10.87\%} & 4.35\% & 8.70\% & 8.70\% & \textbf{15.22\%} \\
			Standard controller for segmentation images & 2.17\% & 2.17\% & 6.52\% & 21.74\% & 36.36\% \\
			\rowcolor[rgb]{ .851,  .851,  .851} Transferred controller for segmentation images & 2.17\% & 2.17\% & \textbf{5.43\%} & \textbf{17.39\%} & \textbf{31.82\%} \\
			\bottomrule
		\end{tabular}%
		\label{tab:addlabel}%
		\begin{tablenotes}
			\footnotesize
			\item[1] Bold values are winning results
		\end{tablenotes}
		\label{tab:addlabel}%
	\end{table*}%

	We conducted the experiment as illustrated in Fig. 8 to evaluate the knowledge-transfer ability in FIL. As presented in Fig. 9, we compare the six models in a neighborhood environment. Corresponding to every type of training data, we obtained a pair of policies: a transferred policy and a general policy. So, there are three pairs of policies generated in the experiment. The performance of controllers based on these policies in key challenging tasks are presented in Fig. 10. The results are summarized in Table 2. From the results, we can see that the imitation learning models obtained in cloud robotic system perform significantly better in accuracy, compared with general models that trained by traditional imitation learning without shared knowledge. 
	FIL improves the training process of imitation learning with the help of shared knowledge. There is a pre-trained model from the cloud for transfer in local imitation learning. So, there is no need for local robots to learn from scratch. We present the comparison of train process in Fig 10. From the figure, we can see that the transferred policies have lower error starting point and the error value.
	\begin{figure*}[thpb]
		\centering
		\includegraphics[width=0.8\linewidth]{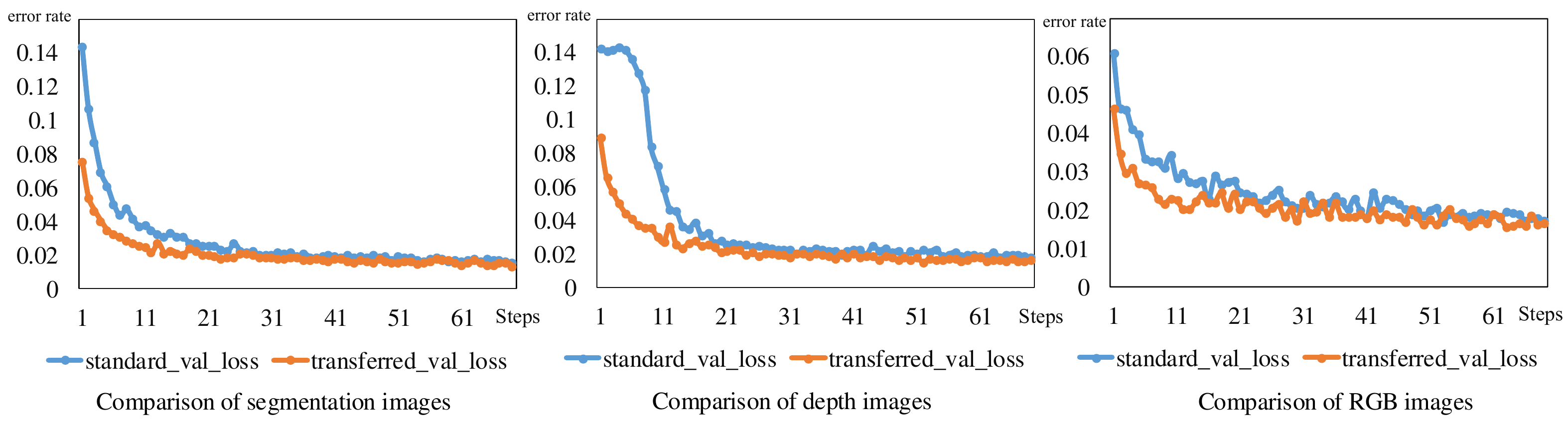}
		\caption{Training process comparison of general learning and transfer learning in FIL}
		\label{fig:architecture}
	\end{figure*}
	\begin{figure}[thpb]
		\centering
		\subfigure[Rain]{
			\label{fig:subfig:a} 
			\includegraphics[width=0.77in]{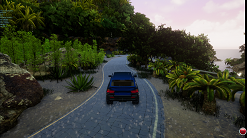}}
		\subfigure[Snow]{
			\label{fig:subfig:b} 
			\includegraphics[width=0.77in]{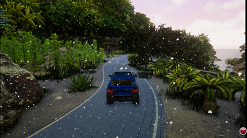}}
		\subfigure[Fog]{
			\label{fig:subfig:a} 
			\includegraphics[width=0.77in]{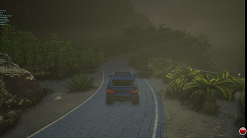}}
		\subfigure[Dust]{
			\label{fig:subfig:b} 
			\includegraphics[width=0.77in]{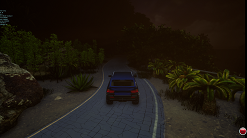}}
		\caption{Bad weather conditions, rain, snow, fog, sand and dust. We conducted data collection and comparison experiments in these environments}
		\label{fig:subfig} 
	\end{figure}
	Local policy models transferred by FIL also have better generalization. Controllers based on transferred policies from FIL perform better in different weather compared with policies trained by general imitation learning. As presented in Fig.12, we conducted the experiments for controllers in different weathers. The results are presented in the last three rows of the Table 2. The results showed that the model from FIL could improve the accuracy of the controller from general imitation learning in bad weather. 
	\section{conclusion}
	In this work, we propose an imitation learning framework for cloud robotic systems with heterogeneous sensor data sharing, named Federated Imitation Learning (FIL). FIL is capable of improving imitation learning efficiency and accuracy of local robots by taking advantage of knowledge from other robots in the cloud robotic system. Additionally, we propose the knowledge fusion algorithm and introduce a transfer method in FIL. Our approach is able to fuse heterogeneous knowledge of local robots. Finally, the framework and algorithms are verified in a self-driving task.
	
	For future work, we expect to research the scalability problem of the platform. How to scale FIL if there are more autonomous cars or more types of sensor data? Although FIL is capable of dealing this issue by simpily expanding the cloud dataset, further work on convergence  justification of the fusion process is needed.
	
\end{document}